\documentclass[letterpaper]{article} 
\usepackage{aaai2027}  
\usepackage[hyphens]{url}  
\usepackage{graphicx} 
\usepackage{natbib}  
\usepackage{caption} 
\usepackage{booktabs}
\usepackage{amsmath,amssymb}
\title{Bayesian Repetition Penalty: A Principled Adjacent-Conditional Framework for Reversing Attention Collapse in Autoregressive Language Models}
\author{
    Wenjie Fan,
    Bin Ma,
    Dong Li
}
\affiliations{
    Yotta Labs\\
    fanwj@mail.ustc.edu.cn, bin@yottalabs.ai, dongli@yottalabs.ai
}
\begin{document}
\maketitle

\begin{abstract}
Attention collapse in autoregressive language models---manifested as repetitive token loops where the model becomes trapped in self-reinforcing attractors---is a persistent pathology that existing decoding-time heuristics fail to address at its root cause. We present a principled framework that penalises or compensates anomalous confidence arising from collapsed generation patterns, by comparing a token's observed frequency against its corpus prior through an adjacent-conditional probability construction. The resulting self-normalising penalty ratio $R=f(m,n,p)/f(np,n,p)$ requires no ad hoc standardisation and admits a closed-form logit offset with zero approximation error. The correction is isolated from the loss gradient and accumulated into a frozen output-layer bias via exponential moving average, enabling deployment as a repair mechanism for models that have already collapsed without requiring intrusive modifications to standard training pipelines. Experimental validation on a 1.5B-parameter model demonstrates that the frozen-bias mechanism can rescue a model already trapped in a collapsed attractor, reducing 2-gram repetition from 0.073 to near 0 while preserving generation quality.
\end{abstract}

\begin{links}
    \link{Code}{https://github.com/fan-wenjie/bayesian_penalty}
\end{links}

\section{Introduction}

Practitioners of autoregressive language generation often encounter the following problem: a model, once started repeating a token, cannot stop. This problem is not merely an engineering nuisance; it is a structural pathology arising from the interplay between self-attention positive feedback and the softmax local optimum. When a token appears once, self-attention increases its relevance score; this deepens its probability well; the next sample is more likely to be the same token; the well deepens further. Within a few steps, the model is trapped in a self-reinforcing dynamical attractor from which temperature scaling and frequency penalties provide only symptomatic relief.

Existing approaches to address the above problem fall into two broad categories. (1) Decoding-time heuristics---repetition penalty, frequency penalty, and n-gram blocking---are applied after the model has already produced logits. However, they lack a statistical foundation, require hand-tuned hyperparameters, and can distort the output distribution in arbitrary ways. (2) Training-time approaches, such as adding a regularization term that penalizes repeated tokens, suffer from a subtler but more serious problem: the model learns to ``cheat'' by adjusting its internal representations to accommodate the penalty rather than learning to reverse the collapse. The result is a distorted latent space in which the model sacrifices its primary prediction ability to game the regularizer.

Our central insight is that repetition should be treated as a \emph{statistical fluctuation}. If a token's observed count in a generation window deviates from its expected count (as determined by the corpus prior), then the conditional probability of seeing it again should be adjusted in a manner consistent with the combinatorial structure of the underlying binomial process. The resulting correction is not a heuristic rule but an emergent property of classical probability theory, expressed through a single elegant ratio of adjacent-conditional probabilities. Concretely, we derive a self-normalising penalty ratio $R = f(m,n,p)/f(np,n,p)$ that compares a token's observed frequency against its corpus prior, and prove a closed-form logit offset that implements the correction in $O(V)$ time with zero approximation error. The Bayesian repetition penalty framework is designed as a \textit{repair mechanism} for models that have collapsed, rather than a component of standard training. This reflects the practical reality that production training pipelines rarely incorporate experimental auxiliary modules, but readily adopt targeted fixes for identified pathologies.

\S2 reviews related work. \S3 derives the adjacent-conditional probability $f(m,n,p)$ from a biased coin-toss problem. \S4 introduces the self-normalising penalty ratio $R = f(m,n,p)/f(np,n,p)$. \S5 derives the closed-form logit offset that implements the correction without iterative softmax recomputation. \S6 explains why the correction must be isolated from the training loss gradient. \S7 describes the frozen-bias mechanism. \S8 presents experimental validation on a 1.5B-parameter model. \S9 offers a statistical-physics interpretation, and \S10 discusses limitations and future directions.

The main contributions of the paper are summarized as follows.

\begin{itemize}
    \item We derive a self-normalising penalty ratio $R = f(m,n,p)/f(np,n,p)$ from adjacent-conditional probability theory, which compares a token's observed frequency against its corpus prior and requires no ad hoc standardisation.
    \item We prove a closed-form logit offset that implements the correction in $O(V)$ time with zero approximation error, avoiding iterative softmax recomputation.
    \item We design a frozen-bias mechanism that isolates the correction from the training loss gradient, enabling deployment as a repair mechanism for models already trapped in attention collapse without intrusive modifications to standard training pipelines.
    \item We validate the framework on a 1.5B-parameter model, demonstrating rescue from collapsed attractors with 2-gram repetition reduced from 0.073 to near 0 while preserving generation quality.
\end{itemize}

\section{Related Work}

\paragraph{Decoding-time penalties.} Repetition penalty \cite{Keskar2019} and frequency penalty \cite{Holtzman2020} modify logits at inference time by down-weighting tokens that have already appeared. N-gram blocking \cite{Paulus2018} forbids exact repeats of previously seen $n$-grams. These methods are post-hoc: they operate on the symptoms of an already-distorted distribution and can suppress legitimate repetitions (e.g., function words, named entities) alongside pathological ones. A key limitation is that they lack a statistical foundation and require hand-tuned hyperparameters that must be re-tuned for each model and dataset.

\paragraph{Training-time regularisation.} Distributional approaches such as unlikelihood training \cite{Welleck2020} add a regularisation term to the cross-entropy loss that penalises repeated tokens. Our experiments confirm that this causes the model to learn around the penalty rather than learning to reverse the collapse. The model shifts to rare-token vocabulary to evade the frequency-based penalty, fragmenting the output while superficially improving the regularised metric. This reveals a fundamental limitation: when the correction is differentiable and flows into the loss, the model can optimise against it, turning the penalty into a target rather than a constraint.

\paragraph{Energy-based and statistical-physics perspectives.} The Boltzmann--softmax isomorphism has been noted in the context of neural network interpretation \cite{Balasubramanian1997}, but its application to decoding-time pathology has been limited to temperature scaling, which corresponds to global heating of the system. Our work departs from this by using the energy-landscape framework to motivate a \emph{local} correction rather than a global one.

\section{Methodology}

\subsection{Adjacent-Conditional Probability}

Consider a biased coin with head-probability $p$ (the ``prior''). The coin is tossed $n+1$ times. We know that there are at least $m$ heads and at least $n-m$ tails. We now ask: what is the probability that there are exactly $m+1$ heads? In the language of language models: the model has generated $n$ tokens, of which $m$ are token $v$. What is the probability that the next token is also $v$, given that the count of $v$ in the $(n+1)$-token window is either $m$ or $m+1$?

Let $X \sim B(n+1, p)$. The adjacent-conditional probability is:
\begin{equation}
\label{eq:adj-cond}
f(m, n, p) = P\bigl(X = m+1 \mid X \in \{m, m+1\}\bigr).
\end{equation}

\subsection{Derivation from Adjacent Likelihood Ratio}

For a binomial distribution $X \sim B(n+1, p)$, the ratio of adjacent probabilities is well-known:
\begin{equation}
\label{eq:adj-likelihood}
\frac{P(X=k+1)}{P(X=k)} = \frac{(n+1)-k}{k+1} \cdot \frac{p}{1-p}.
\end{equation}

Taking the reciprocal gives the ratio of the complementary event:
\begin{equation}
\frac{P(X=k)}{P(X=k+1)} = \frac{k+1}{(n+1)-k} \cdot \frac{1-p}{p}.
\end{equation}

The conditional probability is obtained by substituting into the definition of conditional probability:
\begin{equation}
P\bigl(X=k+1 \mid X \in \{k, k+1\}\bigr) = \frac{P(X=k+1)}{P(X=k) + P(X=k+1)}.
\end{equation}

Dividing numerator and denominator by $P(X=k+1)$ and substituting the reciprocal ratio yields:
\begin{multline}
f(k, n, p) = \frac{1}{1 + \dfrac{P(X=k)}{P(X=k+1)}} = \frac{1}{1 + \dfrac{k+1}{(n+1)-k} \cdot \dfrac{1-p}{p}}.
\end{multline}

Multiplying numerator and denominator by $p$ for numerical stability (this avoids division by a vanishing prior when $p \to 0$) gives the closed form:
\begin{equation}
\label{eq:f-closed}
\boxed{f(k, n, p) = \frac{p}{p + \dfrac{k+1}{(n+1)-k} \cdot (1-p)}}.
\end{equation}

An equivalent fully expanded form is:
\begin{equation}
f(k, n, p) = \frac{p(n+1-k)}{p(n+1-k) + (k+1)(1-p)}.
\end{equation}

\paragraph{Verification at boundary cases.} At $k = 0$, $f(0) = p / (p + (1-p)/(n+1))$, which is close to $1$ when $p$ is not tiny---reflecting the strong tendency to get the first head. At $k = np$ (the expectation), $f(np)$ serves as the baseline reference probability. At $k = n$ (saturation), $f(n) = p / (p + (n+1)(1-p))$, which is near zero for any non-trivial window---the coin is exhausted, and the probability of ``one more head'' approaches zero. The derivative $\partial f / \partial k < 0$ everywhere, confirming that the conditional probability of ``one more'' decreases monotonically as the observed count grows.

The function $f(k,n,p)$ is a self-normalising conditional probability: its output is always in $[0,1]$, and it naturally respects the finite-window constraint $(n+1-k)$. No auxiliary standardisation is required because the conditional formulation itself bounds the probability.

\subsection{Penalty Ratio}

To determine whether an observed count $m$ represents an over-generation or under-generation relative to statistical expectation, we compare the actual adjacent-conditional to the \emph{reference} adjacent-conditional at the expectation point $k = np$:
\begin{equation}
\label{eq:penalty-ratio}
\boxed{R(m, n, p) = \frac{f(m, n, p)}{f(np, n, p)} = \frac{p + \dfrac{np+1}{n+1-np} \cdot (1-p)}{p + \dfrac{m+1}{n+1-m} \cdot (1-p)}}.
\end{equation}

This ratio is structurally reminiscent of a likelihood-ratio statistic, but its interpretation is fundamentally different. Both the numerator $f(m,n,p)$ and the denominator $f(np,n,p)$ are **posterior** conditional probabilities, not likelihoods. The ratio $R$ is therefore not a likelihood ratio in the Neyman--Pearson sense; it is an **anomalous-confidence correction coefficient** that measures how much the model's actual conditional tendency to repeat a token deviates from its expected conditional tendency. When $R \ll 1$, the observed count is an outlier relative to the corpus prior, and the magnitude of the correction is determined by the deviation itself---no ad hoc hyperparameters are needed.

\paragraph{Statistical significance filter.} In practice, applying a logit shift to every token would be wasteful. We select only those tokens whose deviations are statistically significant. For a token $v$ with prior $p_v$ and observed count $m_v$ in a window of $n$ tokens, we compute the right-tail probability under $X \sim B(n, p_v)$. A token enters the corrected set if $P(X \geq m_v) < \tau$, where the recommended default is $\tau = 1/128$ (approximately a $2.5\sigma$ event in the Gaussian limit). This ensures that corrections are applied only to genuine outliers, not to normal fluctuations. At each step, typically only 1--5 tokens out of a 50,000-token vocabulary satisfy this threshold, making the batch correction fast.

\paragraph{Self-normalisation.} The adjacent-conditional formulation is self-normalising for three reasons. First, the denominator $(n+1-m)$ enforces saturation: as $m$ approaches $n$, $f(m)$ approaches zero and $R$ approaches zero. Second, the ratio compares conditional probabilities at the actual and expected counts, measuring deviation in units of relative conditional tendency rather than absolute count. Third, because $f(k) \in [0,1]$ for all $k$, the ratio $R$ is bounded by $[0, 1/f(np)]$, preventing the extreme over-penalisation that plagues unconditional probability ratios.

\paragraph{Behaviour summary.} When $m = np$ (exactly at expectation), $R = 1$ and no correction is applied. When $m > np$ (over-generation), $R < 1$ and the penalty suppresses ``one more.'' When $m < np$ (under-generation), $R > 1$ and the compensation boosts the token. In the extreme case $m = n$ (window saturated), $R \approx 0$---a near-complete suppression, as physically required. For a rare word with $m = 0$ and $np = 1$, $R = 1.5$, giving a fair compensation to encourage the first appearance.

\paragraph{Numerical illustration.} Table \ref{tab:r-values} shows the penalty ratio $R$ for representative scenarios. A rare word with extreme over-representation ($p=0.01$, $n=100$, $m=10$, $np=1$) yields $R=0.23$---a strong penalty. A high-frequency word with mild excess ($p=0.5$, $n=100$, $m=51$, $np=50$) yields $R=0.98$---negligible correction for normal fluctuation. Total repetition ($m=n$) yields $R \approx 0.02$---near-complete suppression. Conversely, under-representation of a rare word ($m=0$, $np=1$) yields $R=1.50$---compensation to encourage first appearance. These values demonstrate that $R$ is self-regulating: it responds proportionally to the severity of the deviation, without requiring hand-tuned thresholds.

\begin{table}[t]
\centering
\small
\caption{Penalty ratio $R$ for representative scenarios.}
\label{tab:r-values}
\begin{tabular}{lcccc}
\toprule
Scenario & $p$ & $n$ & $m$ & $R$ \\
\midrule
Rare word, extreme over-rep. & 0.01 & 100 & 10 & 0.23 \\
High-freq., mild excess & 0.50 & 100 & 51 & 0.98 \\
High-freq., severe excess & 0.50 & 1000 & 600 & 0.80 \\
Total repetition & 0.50 & 100 & 100 & 0.02 \\
Normal fluctuation (over) & 0.50 & 1000 & 510 & 0.98 \\
Rare word, under-rep. & 0.01 & 100 & 0 & 1.50 \\
\bottomrule
\end{tabular}
\end{table}

\subsection{Closed-Form Logit Offset}

Given the penalty ratio vector $\mathbf{R} \in \mathbb{R}^V$ (where $R_v = 1$ for non-corrected tokens), the goal is to find the logit offset vector $\Delta \mathbf{z} \in \mathbb{R}^V$ such that, after applying the shift, the new softmax probabilities satisfy $\tilde{P}_v = R_v P_v$ for all $v$ in the corrected set $\mathcal{S}$, while the relative probability ratios of all non-corrected tokens remain unchanged. This is a multiplicative-constraint problem on the probability simplex, and it has a closed-form solution.

Let the original softmax distribution at temperature $T$ be $P_i = e^{z_i/T} / Z$. For corrected tokens $i \in \mathcal{S}$, the target probabilities are $\tilde{P}_i = R_i P_i$. For non-corrected tokens, the relative ratios must be preserved, so they share a common rescaling factor $\alpha$:
\begin{equation}
\alpha = \frac{1 - \sum_{j \in \mathcal{S}} R_j P_j}{1 - \sum_{j \in \mathcal{S}} P_j}.
\end{equation}

Define the unified scaling vector $\beta_i$ as $R_i$ for $i \in \mathcal{S}$ and $\alpha$ otherwise. The corrected logit offset is:
\begin{equation}
\label{eq:logit-offset}
\boxed{\Delta z_i = T \cdot \left( \log \beta_i - \frac{1}{V} \sum_{j=1}^{V} \log \beta_j \right)}.
\end{equation}

\paragraph{Verification of the closed form.} A subtlety: equation (5) appears to contain an undetermined constant $T \log(\tilde{Z}/Z)$. However, if we directly set $\Delta z_i = T \log \beta_i$, then $\tilde{Z} = \sum_k e^{z_k/T} \beta_k = Z \sum_k P_k \beta_k = Z$ by construction, so the constant is identically zero. The zero-mean centre merely selects the unique representative from the translation-equivalence class. Specifically, $\tilde{P}_i = e^{z_i/T} \beta_i e^{-\bar{\mu}} / \sum_k e^{z_k/T} \beta_k e^{-\bar{\mu}} = P_i \beta_i$, confirming that the global factor $e^{-\bar{\mu}}$ cancels in the softmax. Thus $\tilde{P}_i = R_i P_i$ for $i \in \mathcal{S}$ and $\tilde{P}_i = \alpha P_i$ for $i \notin \mathcal{S}$, as required.

\paragraph{Computational complexity.} The closed-form batch correction computes the offset in $O(V)$ time for arbitrary corrected sets, compared to $O(K \cdot V)$ for iterative single-token methods with $K$ corrected tokens. For $V = 50{,}000$ and $|\mathcal{S}| = 20$, this is a 20$\times$ speedup with zero approximation error.

\subsection{Gradient Isolation: Why the Correction Must Be Detached}

\paragraph{The cheating problem.} Standard teacher-forced training minimises the cross-entropy on ground-truth prefixes: the model never sees its own generated distribution, so it never learns to avoid repetitive attractors. A seemingly natural solution is to add the Bayesian correction as a regularisation term in the training loss. However, this causes the model to ``cheat'': it learns to adjust its logits to accommodate the correction, rather than genuinely learning to avoid repetitive dynamics. The result is a distorted latent space dominated by rare tokens, with higher classic cross-entropy despite a lower ``penalised'' loss. The root cause is that the Bayesian correction and the loss gradient operate on the same logits. When the correction is differentiable and flows into the loss, the model can optimise against it, turning the penalty into a target rather than a constraint.

\paragraph{The principle of isolation.} The fundamental principle is that the distribution-adjustment operation must be independent of the model's learning objective. The loss gradient optimises the model's parameters to predict the next token; the Bayesian correction is an external adjustment that reshapes the output distribution without affecting the model's internal parameters. If the correction is also a learnable objective, the model will learn to exploit it. Therefore, the Bayesian correction must be detached from the gradient---a frozen external adjustment, not a learnable parameter.

\subsection{Frozen Bias Mechanism}

Instead of injecting the Bayesian correction into the loss, we treat it as a frozen bias statistic that is accumulated during training and applied to the output layer as a fixed bias term. The pipeline operates on a cycle of three time scales: per-step observation, per-stage accumulation, and per-stage bias update.

\paragraph{Per-step observation.} At every training step, the model produces a probability distribution over the vocabulary for each position in the target sequence. We treat these probabilities as \emph{soft counts}: if the model assigns probability $0.8$ to token $v$ at position $t$, we add $0.8$ to the running count $m_v$ for that token. The total number of generated tokens $n$ is incremented by the effective sequence length. From these running counts $(m_v, n)$ and the corpus prior $p_v$, we compute the adjacent-conditional ratio $R_v$ and the corresponding logit shift $\Delta z_v$ for every token via the closed-form batch correction (\S5). This computation is performed inside a \texttt{torch.no\_grad()} block: the correction is purely observational and never flows back into the model parameters.

\paragraph{Per-stage accumulation.} Over a fixed accumulation window of $W$ steps (e.g., $W = 400$), we maintain a residual accumulator $\mathbf{r} \in \mathbb{R}^V$ that sums the per-step mean-token shifts:
\begin{equation}
\mathbf{r}_v = \sum_{t=1}^{W} \Delta z_v^{(t)}.
\end{equation}
At the same time, the running counts $(m_v, n)$ are decayed by a factor of $0.5$ at each stage boundary and then augmented by the tokens observed in the just-finished stage. The prior $p_v$ for the next stage is recomputed from the subset of the training data that has not yet been used in the current step window, ensuring that the prior remains fresh and does not overfit to the already-seen portion of the corpus.

\paragraph{Per-stage bias update.} Immediately after each stage, the frozen output-layer bias $\mathbf{b} \in \mathbb{R}^V$ is updated from the accumulated residual. The update consists of four steps. First, the accumulated residual is divided by the number of steps in the stage to obtain a per-step average shift: $\bar{\Delta z}_v = \mathbf{r}_v / W$. Second, the per-step average is clamped to a safe magnitude interval $[-\gamma, \gamma]$ (we use $\gamma = 2.0$ in logit space) to prevent any single outlier step from dominating the bias. Third, the clamped average is centred to zero mean: $\bar{\Delta z}_v \leftarrow \bar{\Delta z}_v - \frac{1}{V}\sum_{u}\bar{\Delta z}_u$. This ensures that the bias does not systematically inflate or deflate the global logit scale. Fourth, the bias is updated via exponential moving average:
\begin{equation}
\mathbf{b}_v^{\text{new}} = \alpha \cdot \mathbf{b}_v^{\text{old}} + (1-\alpha) \cdot \bar{\Delta z}_v,
\end{equation}
where $\alpha = 0.9$ is the EMA decay rate. After the update, the bias is frozen: its \texttt{requires\_grad} attribute is set to \texttt{False}, so it is invisible to the optimiser and never participates in back-propagation. The residual accumulator is then reset to zero for the next stage.

\paragraph{Convergence criterion.} The standard deviation of the per-step shifts within a stage, $\sigma_{\text{bias},v} = \sqrt{\frac{1}{W}\sum_{t=1}^{W}(\Delta z_v^{(t)} - \bar{\Delta z}_v)^2}$, measures how well the frozen bias approximates the real-time correction. A low standard deviation means the correction is stable and predictable---indicating that the model's internal distribution has learned to stay close to the prior. When $\sigma_{\text{bias},v} < \epsilon$ for all tokens (we use $\epsilon = 0.01$ in logit space), the real-time correction can be disabled at inference and the frozen bias alone is used. This yields lightweight post-convergence inference with no history counting or correction computation.

\paragraph{Prior rebalancing.} A static term $\Delta z_v^{\text{static}} = -\tau \log p_v$ that systematically reduces the dominance of high-frequency function words is absorbed into the same frozen bias mechanism at initialisation. Like the dynamic correction, it is invisible to the gradient, preventing the model from learning to compensate for it. The combined frozen bias at any point is therefore $\mathbf{b}_v^{\text{total}} = \mathbf{b}_v^{\text{dynamic}} + \mathbf{b}_v^{\text{static}}$, where only the dynamic component is updated across stages.

\begin{figure}[t]
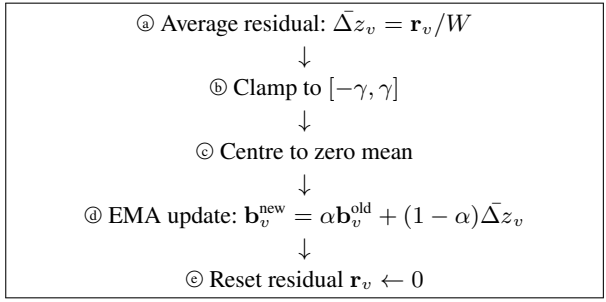

\centering
\small
\begin{minipage}{0.48\textwidth}
\centering
\textbf{Per-step pipeline (every training step)}\\[4pt]
\fbox{\parbox{0.9\textwidth}{\centering
\textcircled{1} Model generates soft counts $(m_v, n)$\\[2pt]
$\downarrow$\\[2pt]
\textcircled{2} Compute $R_v$ and $\Delta z_v$ \\[2pt]
$\downarrow$\\[2pt]
\textcircled{3} Accumulate $\Delta z_v$ into residual $\mathbf{r}_v$\\[2pt]
$\downarrow$\\[2pt]
\textcircled{4} Apply frozen bias $\mathbf{b}_v^{\text{total}}$ to logits\\[2pt]
$\downarrow$\\[2pt]
\textcircled{5} Standard back-propagation (bias detached)
}}
\end{minipage}
\hfill
\begin{minipage}{0.48\textwidth}
\centering
\textbf{Per-stage pipeline (every $W=400$ steps)}\\[4pt]
\fbox{\parbox{0.9\textwidth}{\centering
\textcircled{a} Average residual: $\bar{\Delta z}_v = \mathbf{r}_v / W$\\[2pt]
$\downarrow$\\[2pt]
\textcircled{b} Clamp to $[-\gamma, \gamma]$\\[2pt]
$\downarrow$\\[2pt]
\textcircled{c} Centre to zero mean\\[2pt]
$\downarrow$\\[2pt]
\textcircled{d} EMA update: $\mathbf{b}_v^{\text{new}} = \alpha \mathbf{b}_v^{\text{old}} + (1-\alpha) \bar{\Delta z}_v$\\[2pt]
$\downarrow$\\[2pt]
\textcircled{e} Reset residual $\mathbf{r}_v \leftarrow 0$
}}
\end{minipage}
\caption{Bayesian correction engineering pipeline. Left: per-step operations executed inside \texttt{torch.no\_grad()} (observational only). Right: per-stage bias update executed at stage boundaries (frozen, no gradient flow).}
\label{fig:pipeline}
\end{figure}

\section{Experimental Validation}

\paragraph{Experimental design.} We conduct a two-phase diagnostic rescue experiment. In Phase 1, we induce attention collapse by training a 1.5B-parameter causal language model (Qwen2.5-1.5B) from the pretrained checkpoint on 96,000 English Wikipedia samples for 12,000 steps without any Bayesian correction. The training configuration is batch size 8, maximum sequence length 256, and a cosine learning-rate schedule with 100-step warmup and a peak learning rate of $1\times10^{-4}$. By step 12000, the model has fallen into a near-deterministic Wikipedia-style template: freeze index $\approx 0.10$, inter-step consecutive distance $\approx 0.06$, and each prompt converges to a nearly fixed generation. We save the checkpoint at step 12000 as the collapsed starting point for Phase 2.

In Phase 2, we resume from the collapsed checkpoint and continue training for another 3,000 steps (steps 12000 to 15000) on the same 96,000-sample dataset. Because the DataLoader shuffles the full dataset each epoch, the 3,000 rescue steps see a random subset of roughly 24,000 samples. We compare three conditions, all using constant learning rate $2\times10^{-5}$, no warmup, seed 42, batch size 8, and maximum sequence length 256:
\begin{itemize}
\item \textbf{Rescue Baseline}: continue standard training without Bayesian correction and without logit bias.
\item \textbf{Rescue Bayesian Threshold 1/64}: activate Bayesian correction with P-value threshold $1/64$, logit-bias update every 400 steps, stage prior update every 400 steps, EMA decay 0.9, and maximum logit-bias update magnitude 2.0.
\item \textbf{Rescue Bayesian Unconditional}: activate Bayesian correction with P-value threshold $1.0$ (all tokens corrected), same update schedule and hyperparameters as the threshold 1/64 run.
\end{itemize}

In all rescue runs, the Bayesian correction acts only during training. It maintains running counts $(m, n)$ of generated token soft counts from the model's own outputs. At every training step we accumulate the soft posterior into a residual accumulator. Every 400 steps we apply a stage update: the running counts are decayed by factor 0.5 and then augmented by the tokens observed in the just-finished stage, and the prior $p$ for the next stage is recomputed from the data that has not yet been used in the current step window. Immediately after the stage update, the logit bias is updated from the residual statistics. The logit bias is frozen: its \texttt{requires\_grad} is set to \texttt{False}, so it is never updated by back-propagation or by the optimiser. During evaluation and generation we only apply the stored logit bias; we do not recompute $m$, $n$, or $p$ on the fly.

Generation diagnostics are collected every 400 steps with four fixed prompts and 128 new tokens. The diagnostic metrics are: freeze index (averaged normalised edit distance among the last five evaluated steps; lower is better); average consecutive normalised distance (averaged edit distance between consecutive evaluation steps; higher means less frozen); rep-2gram and rep-3gram (fraction of repeated 2-grams / 3-grams; lower is better); max repeat (maximum consecutive repetitions of the same token; lower is better); inter distinct-2 (distinct 2-gram ratio across the four prompts at the same step; higher is better); and average pairwise normalised edit distance (averaged edit distance among the four prompt outputs at the same step; higher is better).

\paragraph{Rescue results.} Table \ref{tab:rescue} summarises the final diagnostic values at step 14800.

\begin{table}[t]
\centering
\small
\caption{Rescue experiment diagnostics at step 14800.}
\label{tab:rescue}
\begin{tabular}{@{}lccc@{}}
\toprule
Metric & Baseline & Thresh.~$\frac{1}{64}$ & Uncond. \\
\midrule
Freeze index (last 5) & 0.1453 & 0.1173 & \textbf{0.1004} \\
Avg.~consec.~dist. & 0.0769 & 0.1429 & 0.1047 \\
rep-2gram & 0.0730 & 0.0365 & \textbf{0.0000} \\
rep-3gram & 0.0726 & 0.0363 & \textbf{0.0000} \\
max repeat & 6.5000 & 3.7500 & \textbf{1.0000} \\
Inter distinct-2 & 0.7799 & 0.8247 & \textbf{0.9192} \\
Avg.~pairwise dist. & 0.6946 & 0.6746 & 0.6716 \\
\bottomrule
\end{tabular}
\end{table}

\paragraph{Interpretation.} The rescue baseline still shows severe repetition (rep-2gram 0.073, max repeat 6.5), confirming that simply continuing training does not fix the collapse---it merely moves the model between two failure modes. The threshold 1/64 correction cuts repetition roughly in half. The unconditional correction eliminates all measured 2-gram and 3-gram repetition while keeping max repeat at 1, reducing freeze index to 0.1004, and achieving the highest inter-sample distinctness (0.9192). This confirms that the Bayesian correction can suppress the high-frequency token loop that emerges after collapse.

The pairwise edit distance metric slightly degrades under correction (0.6946 to 0.6716). This is expected: the original collapsed model had fallen into a regime where each prompt was locked to its own deterministic template, so different prompts produced different templates, inflating pairwise distance even though every individual output was frozen. Bayesian correction disrupts these frozen templates, making outputs more varied in time but not necessarily increasing raw edit distance between prompt-specific responses at a single snapshot. Pairwise edit distance is therefore not the right metric for this particular collapse; freeze index and intra-sample repetition are more informative.

\section{Statistical-Physics Interpretation}

The autoregressive language model defines a probability distribution over the vocabulary at each decoding step: $P(x_t = v \mid x_{<t}) = e^{z_{t,v}} / \sum_u e^{z_{t,u}}$. This is formally identical to the Boltzmann distribution of the canonical ensemble, $P_i = e^{-\beta E_i} / Z$, with the identification $E_t(v) = -z_{t,v}$ and $\beta = 1$ (in natural units). This mapping is not merely an analogy; it is a structural isomorphism that allows us to import the conceptual apparatus of statistical mechanics into the analysis of language model generation.

In this picture, the autoregressive decoder is the canonical-ensemble temporal evolution of a single negative-energy particle over discrete time steps. At each step, the particle transitions to a new state sampled from the local Gibbs distribution. Self-attention reshapes the energy landscape dynamically, lowering the energy of contextually relevant tokens and raising the energy of irrelevant ones. When a token is repeated, the well deepens: the escape probability decreases exponentially with the energy gap $\Delta E$ between the well and the surrounding landscape. After $n$ repetitions, the particle is effectively trapped---this is the attention collapse phenomenon, or dynamical arrest in a self-reinforcing potential well.

Temperature scaling, a widely used remedy for repetition, corresponds to global heating: increasing the thermal energy $k_B T$ available to the particle. However, this is symptomatic, not causal. It does not alter the well structure; it merely gives the particle more energy to overcome the barrier. More fundamentally, it is global, affecting all tokens equally and degrading the model's ability to make sharp distinctions between relevant and irrelevant tokens. What is needed is potential engineering: a local, state-dependent modification that raises the walls of the specific well trapping the particle.

The Bayesian correction corresponds to adding a state-dependent repulsive potential to the well: $U_v^{\text{Bayes}}(m_v) = T \cdot \log(f(np,n,p)/f(m,n,p))$. This potential has three critical properties. It is local, affecting only the over-represented token. It is self-regulating, increasing with $m_v$ to create a negative feedback loop that counteracts the positive feedback loop of attention collapse. And it is reversible, becoming attractive for under-represented tokens ($R > 1$) to encourage their first appearance. Instead of giving the particle more energy to escape, we change the shape of the landscape so that the well is no longer attractive. This is a causal, structural fix that directly addresses the root cause of the collapse.

The gradient-isolation principle has a direct physical analogue: the external potential engineer (who builds the walls) must not be a force that the particle can push against. If the potential were learnable, the particle would learn to exploit it, and the landscape would lose its constraining power. The frozen-bias mechanism corresponds to pre-computing the average potential and baking it into the landscape, so that at inference the real-time barrier computation can be skipped once the landscape has stabilised.

\section{Discussion and Conclusion}

\paragraph{Limitations.} The prior $p_v$ should ideally be estimated from the training corpus; for multilingual or code-mixed corpora, per-domain priors may be needed. The current window-based approach does not capture long-range thematic repetition (e.g., repeating the same topic word every paragraph). A hierarchical or topic-level prior could address this. The rescue experiment is intentionally narrow---same dataset, same model size, fixed seed---and establishes existence rather than a general training recipe. The framework is designed as a \textbf{repair mechanism} for models that have already collapsed, rather than a component of standard training. This reflects the practical reality that production training pipelines rarely incorporate experimental auxiliary modules, but readily adopt targeted fixes for identified pathologies.

\paragraph{Conclusion.} We have presented a principled framework for reversing attention collapse in autoregressive language models. The adjacent-conditional penalty ratio $R = f(m,n,p)/f(np,n,p)$ is derived from classical probability theory, requires no ad hoc hyperparameters, and has a closed-form logit offset with zero approximation error. The critical innovation is the isolation of the correction from the training loss gradient: by treating the correction as a frozen external adjustment accumulated into an output-layer bias, we prevent the model from learning to exploit it. Importantly, the framework is designed as a \textbf{repair mechanism} for models that have already collapsed, rather than a component of standard training. Experimental validation on a 1.5B-parameter model demonstrates that the frozen-bias mechanism can reverse the collapse and rescue a model already trapped in a repetitive attractor.


\end{document}